\documentclass{article} 
\usepackage[final]{colm2026_conference}

\usepackage{microtype}
\usepackage{hyperref}
\usepackage{url}
\usepackage{booktabs}

\usepackage{graphicx}
\usepackage{subcaption}
\usepackage{xcolor}
\usepackage[table]{xcolor}
\usepackage{listings}
\usepackage{xspace}
\usepackage{amsmath}
\usepackage{amssymb}
\usepackage{mathtools}
\usepackage{amsthm}
\usepackage{multirow, multicol}

\usepackage[table]{xcolor}
\definecolor{lightgreen}{RGB}{198, 239, 206}
\definecolor{lightred}{RGB}{255, 199, 206}
\newcommand{\good}[1]{\cellcolor{lightgreen}{#1}}
\newcommand{\bad}[1]{\cellcolor{lightred}{#1}}

\usepackage{lineno}

\definecolor{darkblue}{rgb}{0, 0, 0.5}
\hypersetup{colorlinks=true, citecolor=darkblue, linkcolor=darkblue, urlcolor=darkblue}

\newcommand{\metricx}{MetricX-24 QE\xspace}
\newcommand{\cometkiwi}{CometKiwi\xspace}

\title{Penalizing Length: Uncovering Systematic Bias in Quality Estimation Metrics}

\author{
Yilin Zhang$^{1,2}$\thanks{Correspondence to: Yilin
Zhang \textless{}yilinjz@google.com, jasonzh3@andrew.cmu.edu\textgreater{}} \quad
Wenda Xu$^{3}$ \quad
Zhongtao Liu$^{3}$ \quad
Tetsuji Nakagawa$^{2}$ \quad
Markus Freitag$^{3}$ \\
$^{1}$Carnegie Mellon University \quad
$^{2}$Google \quad
$^{3}$Google DeepMind\\
}

\begin{document}

\ifcolmsubmission
\linenumbers
\fi

\maketitle

\begin{abstract}
Quality Estimation (QE) metrics are vital in machine translation for reference-free evaluation and increasingly serve as selection criteria in data filtering and candidate reranking. However, the prevalence and impact of length bias in QE metrics have been underexplored. Through a systematic study of top-performing learned and LLM-as-a-Judge QE metrics across 10 diverse language pairs, we reveal two critical length effects: First, QE metrics consistently over-predict errors with increasing translation length, even for high-quality, error-free texts. Second, when candidates of comparable quality are available for the same source text, learned QE metrics tend to favor shorter translations, whereas LLM-as-a-Judge metrics range from being approximately length-neutral to favoring longer translations. These metric-dependent length effects can favor or penalize translations based on length rather than quality and can propagate into downstream pipelines that rely on QE signals for data selection or system optimization. We trace the root cause of MetricX-24 QE’s cumulative-length penalty to skewed supervision distributions, in which longer error-free examples are underrepresented in training data. As a diagnostic intervention, we apply length normalization during training and show that this simple modification effectively decouples error prediction from sequence length, yielding more reliable QE signals across translations of varying length.
\end{abstract}

\begin{figure*}[h!]
\vspace{-6pt}
\centering
\includegraphics[width=\textwidth]{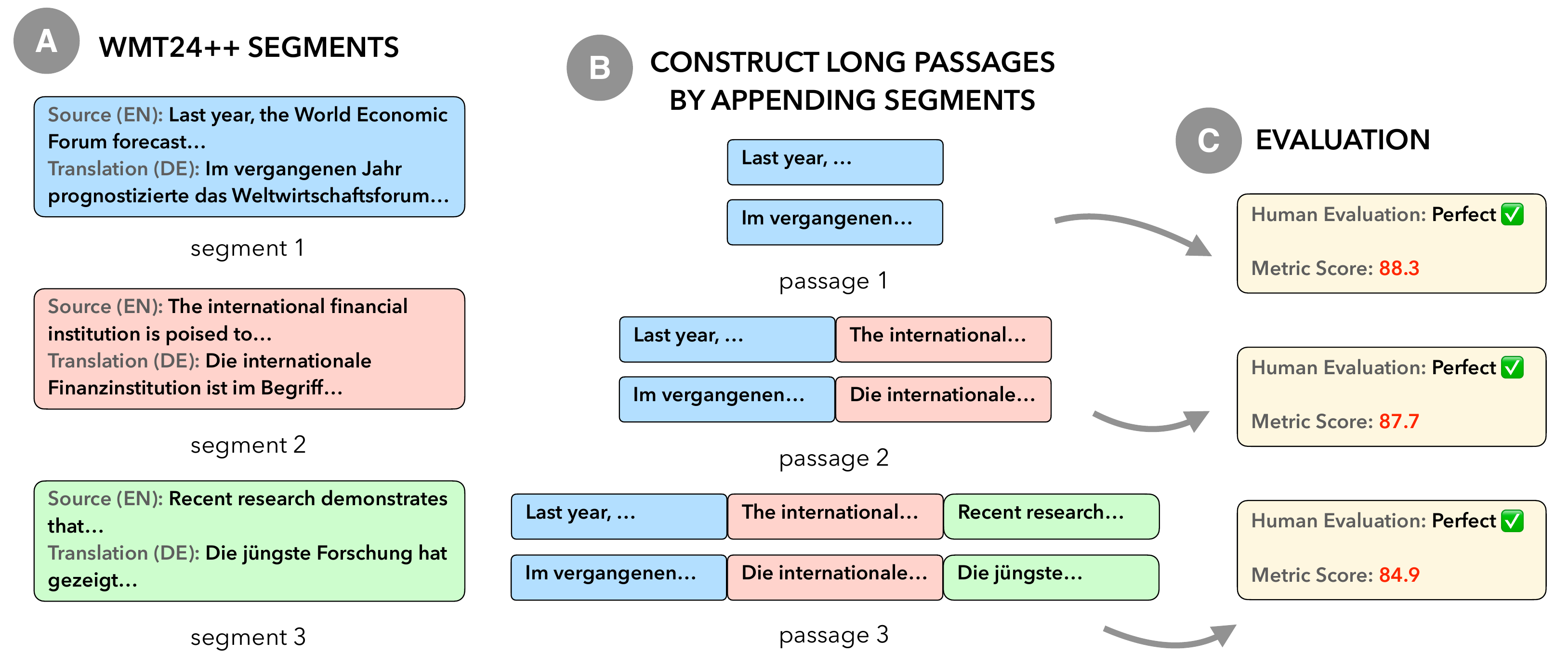}
\caption{QE metrics exhibit a length-related penalty when evaluating long-form translations, as metric scores decrease when translations become longer despite being error-free. As shown, concatenating consecutive gold-standard segments from the same document leads to progressively lower predicted scores, in contrast to human evaluations.}
\label{fig:teaser}
\vspace{-6pt}
\end{figure*}

\section{Introduction}

Quality Estimation (QE) predicts the quality of machine-translated text without relying on reference translations, serving as a vital signal for applications where ground truth is unavailable. Broadly, top-performing QE methods fall into two categories: learned metrics, such as MetricX \citep{juraska-2024-metricx24} and COMET \citep{rei-2020-comet}, which learn to assess the hypothesis directly from annotated data, and off-the-shelf
LLM-as-a-Judge approaches, such as GEMBA \citep{kocmi-2023-GEMBA}, InstructScore \citep{xu-2023-instructscore} and AutoMQM \citep{fernandes-2023-automqm}, which leverage large language models to generate error ratings that approximate human judgments.

Both paradigms ultimately rely on error aggregation schemes rooted in Multidimensional Quality Metrics (MQM) \citep{freitag-2021-mqm}: segment-level quality is computed by linearly accumulating error penalties, and system-level performance is typically obtained by averaging these scores across an evaluation set. While intuitive, this linear aggregation approach introduces several systematic deficiencies. First, as sequence length increases, minor and major errors naturally accumulate, causing long yet high-quality translations to be penalized disproportionately relative to shorter, lower-quality ones. Second, because raw MQM scores are unbounded and scale with text length, longer segments mathematically dominate corpus-level averages, effectively diluting the quality signals of shorter sequences. Prior work has documented related length effects in LLM-as-a-Judge settings \citep{domhan-2025-evaluationtokenseffectinput} and reward models \citep{zhao-2025-biasfitting}, yet the extent to which QE metrics suffer from these failure modes remains largely underexplored. This gap is consequential: length bias can distort system comparisons, impose metric-dependent length preferences in data selection and candidate reranking, and ultimately misguide model development.

In this study, we design two complementary experiments to systematically investigate length bias in QE metrics across 10 language pairs. The first evaluates length invariance by concatenating high-quality, error-free segments into progressively longer translations, assessing whether metric scores remain stable as content accumulates. The second directly compares quality-matched paraphrases of different lengths for the same source. All evaluated metrics assign lower quality as cumulative passages grow (e.g., Figure \ref{fig:teaser}), but the matched-paraphrase experiment reveals model-dependent behavior: learned QE metrics tend to favor shorter hypotheses, whereas LLM-as-a-Judge metrics range from being approximately length-neutral to favoring longer translations.

To probe the root cause of the cumulative-length penalty in a learned metric, we analyze the training data of MetricX-24 QE and find that longer error-free examples are severely underrepresented, inducing a spurious association between sequence length and predicted error. Increasing model capacity does not reliably reduce the effect (\S\ref{analysis:model_size}), indicating that it is not simply a consequence of underfitting. Applying a simple length normalization during training restores alignment with the ground-truth score distribution, confirming that MetricX's bias originates from data skew rather than insufficient model capacity.

\section{Related Work}

\subsection{Length Bias in Evaluation Models}

Learned evaluation metrics have increasingly been shown to be sensitive to input and output sequence length in ways that are unrelated to output quality. In MT evaluation, neural metrics are vulnerable to perturbations such as omissions and repetitions, which often correlate with atypical translation lengths \citep{karpinska-2022-demetr}. More directly, \citet{domhan-2025-evaluationtokenseffectinput} show that longer inputs cause LLM-based MT evaluators to identify fewer error spans and yield less accurate system rankings. Similarly, at the document level, existing MT evaluation methodologies remain constrained to sentence-level assessment due to token limits in metrics and rigid sentence boundaries, motivating segmentation-and-alignment methods for book-length translation evaluation \citep{wang-2025-extending-segale}. These issues are not limited to MT. In LLM-as-a-Judge settings, verbosity bias, the tendency to prefer longer outputs regardless of quality, has been identified as a systematic failure mode \citep{zheng-2023-judging}. RLHF reward models exhibit similar behavior: they inherit length biases from preference data to the point where a simple length-based reward can reproduce much of the observed performance gains \citep{singhal-2024-longwaygo}.

Despite growing evidence that length sensitivity is a recurring issue across learned evaluators, QE metrics, which increasingly serve as selection criteria in data filtering and candidate reranking, have not been systematically compared across learned and LLM-as-a-Judge paradigms along two distinct axes: accumulating correct content and varying verbosity while holding content and quality fixed. We address this gap by auditing length bias across both learned and LLM-as-a-Judge QE metrics.

\subsection{Training Data Artifacts in Learned Metrics}

Biases in learned metrics often originate from imbalances in the training data. For example, \citet{yan-etal-2023-bleurt} identify universal adversarial translations in BLEURT and BARTScore and trace these robustness defects partly to these distributional imbalances. QE models trained on standard datasets tend to rely heavily on target-side fluency rather than accurately assessing fidelity to the source, a pattern that can be traced back to how training examples are constructed \citep{behnke-2022-bias}. Similarly, metrics trained primarily on MT outputs often assign lower scores to human translations, failing to capture their greater lexical and structural diversity \citep{deutsch-2025-wmt24}.

Such biases can arise even earlier, during data collection. LLM-based preference annotators can favor more verbose answers even when quality is comparable, introducing length bias into supervision before model training begins \citep{saito-2023-verbositybias}. More broadly, distributional biases across domains and languages have been widely documented \citep{zouhar-2024-pitfalls}.

Building on this line of work, we focus on length as a specific source of bias in QE. In particular, we analyze MetricX-24 QE as a case study, investigating whether the underrepresentation of long, error-free examples in its training data leads the model to associate longer sequences with higher error rates.

\section{Preliminaries}
\label{preliminary}
\paragraph{Definition of Length Bias.}
Length bias describes a systematic preference for certain output lengths
irrespective of true quality. We formalize this using the standard statistical
definition of bias, conditioned on sequence length. Given a source text $x$
and a model-generated hypothesis $y$, a QE model produces a quality
prediction $\hat{\theta}(x, y)$. Let $l$ denote a length measure of
interest (e.g., the token count of $x$ or $y$). We define the length
bias as:
\begin{equation}
\textrm{Bias}(\hat{\theta}, l) =
\mathbb{E}[\hat{\theta}(x, y)] - \theta
\end{equation}
where $\theta$ denotes the true translation quality, derived from human
assessments following annotation guidelines such as MQM
\citep{freitag-2021-mqm}, and the expectation is taken over all
source--hypothesis pairs $(x, y)$ whose relevant length measure equals
$l$. In our normalized scoring convention, higher scores indicate better
quality. When $\textrm{Bias}(\hat{\theta}, l) > 0$, the model
overestimates quality at length $l$, effectively favoring translations
of that length; when $\textrm{Bias}(\hat{\theta}, l) < 0$, it
underestimates quality, penalizing translations of that length.

In our study, all instances within a comparison group are constructed to
share the same true quality $\theta$. An unbiased QE model should
therefore produce predictions
$\mathbb{E}[\hat{\theta}(x, y)]$ that remain constant
across lengths. A length-biased model, by contrast, yields predictions
that vary systematically with $l$. As we show in subsequent sections,
the dominant pattern is increasingly negative bias as $l$ increases,
indicating that QE metrics systematically penalize longer translations.

\section{Measuring Length Bias in QE Metrics}
\label{sec:experiments}

\subsection{Evaluation Settings}

\paragraph{Task Definition.} 
A QE metric, $M_{\text{QE}}$, assesses the quality of a system's hypothesis $h$ given a source text $x$ without relying on human references, assigning a score based on predefined criteria. We investigate the phenomenon of \emph{length bias} by comparing the scores of groups of source--hypothesis pairs $(x,h)$ with the same human-rated translation quality, where instances in each group are formed by either varying the sequence length of error-free translations or selecting alternative hypotheses of differing lengths for the same source.

\label{experiment_setup}
\paragraph{Dataset.} We use the WMT24++ data \citep{deutsch-2025-wmt24}, which contains human translation and post-edit data for 55 en$\rightarrow$xx language pairs. We focus on a subset of 10 languages---German, Chinese, Spanish, French, Japanese, Korean, Hindi, Arabic, Russian, and Portuguese---selected as the most widely spoken languages in the dataset. WMT24++ translations were produced by professional vendors provided with full document-level context rather than sentence-level segments. As noted by \citet{deutsch-2025-wmt24}, these post-edits often constitute complete rewrites to maximize fluency and accuracy, representing a high-quality gold standard superior to traditional minimal-edit post-editing, and are thus considered error-free. The post-edits also preserve cross-sentential discourse markers and pronominal references, ensuring that the concatenation of contiguous segments reconstructs natural, cohesive paragraphs.

\paragraph{Quality Estimation Metrics.} We evaluate six QE metrics spanning two paradigms: (i) two learned metrics, \metricx \citep{juraska-2024-metricx24} and \cometkiwi \citep{rei-2022-cometkiwi}, and (ii) four AutoMQM variants: a fine-tuned model based on Gemini-2.0 \citep{fernandes-2023-automqm, finkelstein-2024-jacktradesmasterone} and three few-shot-prompted models based on Llama-3.1-8B \citep{llama-2024}, Qwen3-8B \citep{qwen3-2025}, and Gemini-2.5-Pro \citep{gemini-2025}. The few-shot prompt is provided in Appendix~\ref{appendix:automqm_prompt}.

\paragraph{Hypotheses.} We investigate two forms of length bias: (i) bias with respect to source text length and (ii) bias with respect to translation length. For (i), we construct multiple source--translation pairs of equivalent quality to test whether metrics show systematic preferences related to source text length. For (ii), we compare pairs of translations of equal quality generated from the same source, categorize them by relative length, and analyze whether metrics display a consistent preference for a particular category.

\subsection{Bias in Source Text Length}
\label{main-experiment_data-filtering}

\vspace{-2pt}
\begin{figure*}[ht!]
\centering
\begin{tabular}{c}
\includegraphics[width=\textwidth]{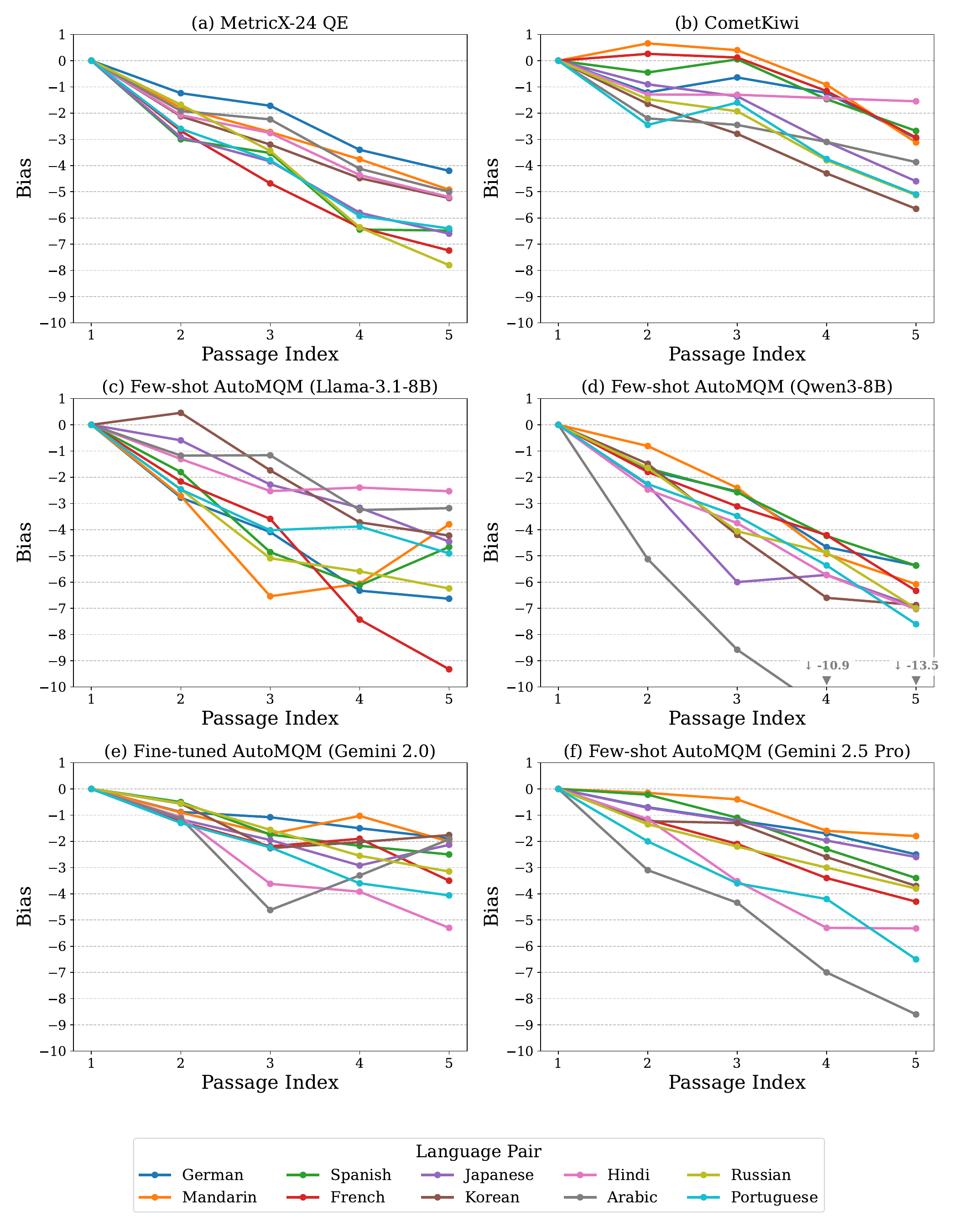}
\end{tabular}
\caption{Metric scores with increasing source text length. Each data point denotes the score change relative to the first passage.}
\label{fig:main-exp_data-filtering}
\end{figure*}
\vspace{-4pt}

\paragraph{Experiment Setup.} For each language pair, we collect the first 5 segments from each document in the WMT24++ data, denoted $s_1, \dots, s_5$, where each segment is approximately 1--2 sentences. We reconstruct the original document flow by gradually appending contiguous segments: for each document $d$, we construct 5 passages $p_1, \dots, p_5$, where $p_i = s_1 + \dots + s_i$. Each passage is a strict prefix of the next ($p_1 \subset p_2 \subset \dots \subset p_5$), ensuring that the resulting sequences represent original, coherent discourse rather than arbitrary concatenations. Since the underlying translations are error-free, appending additional segments can only add correct content; any observed score degradation is therefore attributable to length rather than quality.

This construction has two properties. First, selecting segments from the same document keeps passages within a single domain and minimizes confounds from topic variation. Second, concatenating consecutive segments preserves inter-segment fluency and reduces boundary-induced disfluency.

Since metrics inherit different context-window limits from their base models (e.g., 512 tokens for \metricx), we standardize the context window to the smallest across metrics and discard documents whose first 5 segments exceed 500 tokens using MetricX-24's tokenizer.\footnote{\url{https://huggingface.co/google/mt5-xl}} To control for score variations across language pairs, we normalize scores within each group by subtracting the score of $p_1$ from subsequent passages and report score differences. Figure~\ref{fig:main-exp_data-filtering} presents the resulting trends across language pairs and metrics. When references are available, the bias is attenuated but not eliminated (Appendix~\ref{appendix:reference_based_metrics_length_bias}), confirming that the effect is most pronounced in the reference-free QE setting.

\paragraph{Results.} After orienting all metrics so that higher scores indicate better quality and expressing score changes on a common 0--100 scale, all six metrics assign lower quality to five-segment passages than to one-segment passages across every language pair. Extending the source from 1 to 5 segments results in score drops of 4.0--8.0 points for \metricx, 1.5--5.6 for \cometkiwi, 2.5--9.3 for few-shot AutoMQM (Llama-3.1-8B), 5.4--13.5 for few-shot AutoMQM (Qwen3-8B), 1.7--5.3 for fine-tuned AutoMQM (Gemini 2.0), and 1.8--8.6 for few-shot AutoMQM (Gemini 2.5 Pro). Because all passages are error-free, their true quality $\theta$ is constant; under Section~\ref{preliminary}, an unbiased metric's score change from $p_1$ to $p_i$ should therefore be approximately zero. Appendix~\ref{appendix:proportion_of_decreasing_trends_data_filtering} reports that approximately 80\% of documents exhibit this decreasing trend.

Under the MQM error scheme \citep{freitag-2021-mqm}, penalties of --1 (minor) and --5 (major) on a 25-point scale correspond to --4 and --20 on our 100-point scale. While a marginal decrease might be expected from the cumulative probability of minor imperfections in longer sequences ($P^{N}_{error\text{-}free}$), the observed magnitude, particularly for LLM-as-a-Judge metrics (up to 13.5 points), far exceeds the theoretical decay, indicating a systematic bias against length rather than a valid accumulation of errors.

\subsection{Bias in Error Type and Severity}

\begin{figure*}[t!]
\centering
\begin{tabular}{c}
\includegraphics[width=0.95\textwidth]{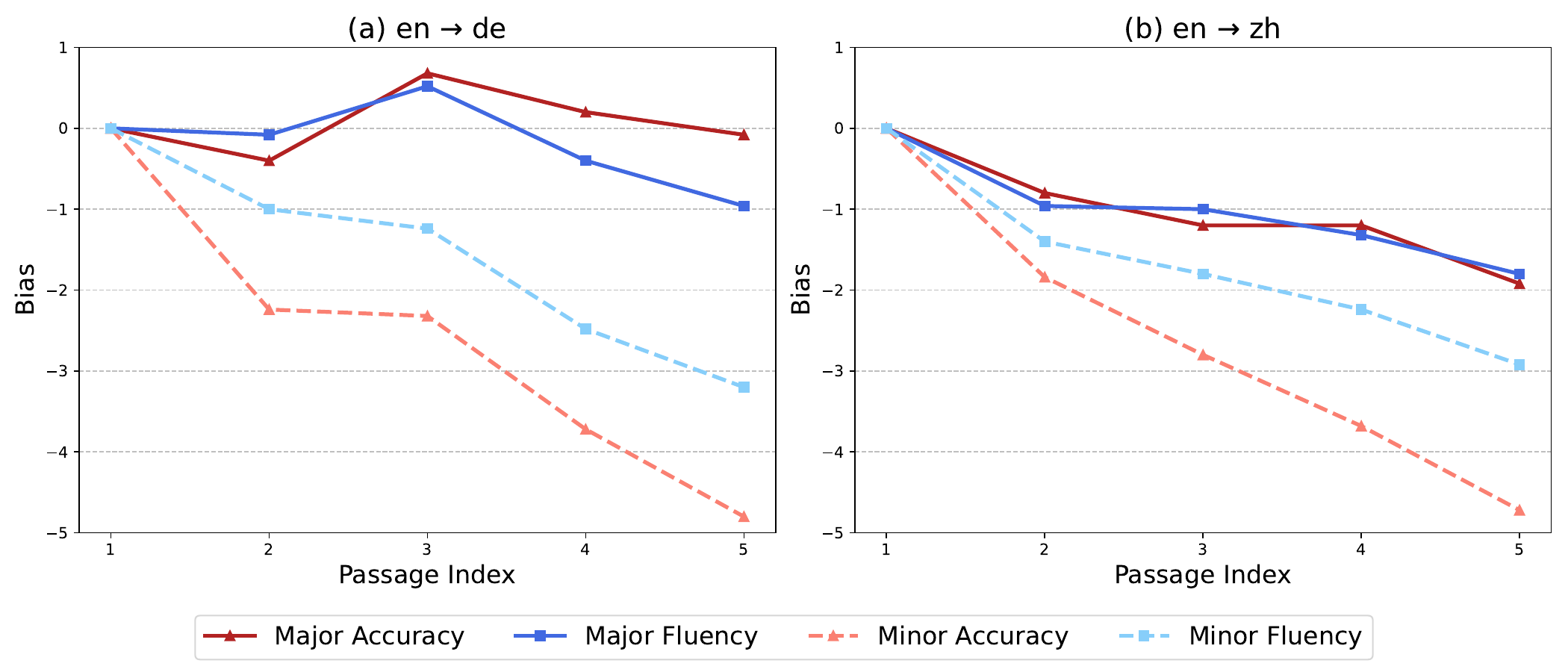}
\end{tabular}
\caption{Trends of \metricx score with different severity/error types, evaluated on WMT24++ en$\rightarrow$de and en$\rightarrow$zh.}
\label{fig:metricx_error-type}
\end{figure*}

\begin{figure*}[t!]
\centering
\begin{tabular}{c}
\includegraphics[width=0.95\textwidth]{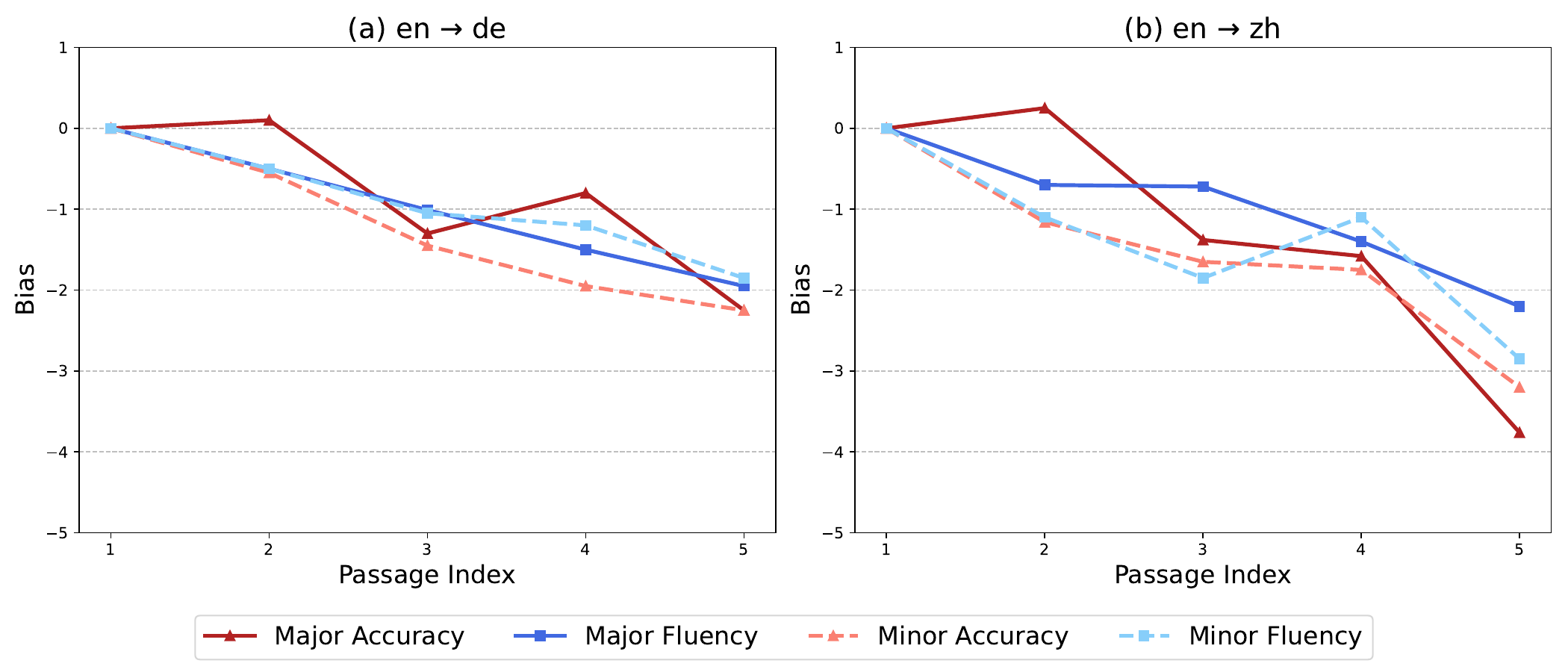}
\end{tabular}
\caption{Trends of fine-tuned AutoMQM score with different severity/error types, evaluated on WMT24++ en$\rightarrow$de and en$\rightarrow$zh.}
\label{fig:gemini-automqm_error-type}
\end{figure*}
\vspace{-6pt}

The preceding experiments establish length bias on error-free translations. We next investigate whether this bias persists when translations contain explicit errors, and whether it interacts with error severity and type. Building on the setup in Section~\ref{main-experiment_data-filtering}, we modify the first segment of each passage by inserting one of four controlled error categories: (a) major accuracy, (b) major fluency, (c) minor accuracy, or (d) minor fluency. Error definitions and few-shot prompts are provided in Appendix~\ref{appendix:error_type_and_severity}. We evaluate using \metricx and fine-tuned AutoMQM and report results for en$\rightarrow$de and en$\rightarrow$zh in Figures~\ref{fig:metricx_error-type} and~\ref{fig:gemini-automqm_error-type}.

For \metricx, inserting a major error in the first segment reduces score fluctuations as additional error-free segments are appended, whereas inserting a minor error has little influence on the existing length bias. Major errors provide a strong signal that outweighs the length effect, delineating a clearer boundary between erroneous and correct content; minor errors contribute substantially weaker cues. For fine-tuned AutoMQM, the results show no consistent interaction between length bias and error severity. Across both metrics, fluency errors have a smaller impact on score changes than accuracy errors, consistent with the observation that metrics are generally more sensitive to accuracy errors, while fluency errors exhibit higher inter-rater disagreement in the training data.

\subsection{Bias in Translation Length}
\label{translation_length_bias}

\paragraph{Experiment Setup.} 
The preceding experiments vary source text length while holding quality
constant. We next test whether metrics also exhibit biased preferences
among translations of different lengths derived from the same source. Here we evaluate \metricx, \cometkiwi, xCOMET-XL \citep{guerreiro-etal-2024-xcomet}, and few-shot AutoMQM based on Llama-3.1-8B, Qwen3-8B, and Gemini-2.5-Flash.

We pre-process source and reference translations from WMT24++ into chunks within a 500-token context window, discarding those shorter than 200 tokens to ensure rephrasing is possible. For each source--reference pair, we prompt Qwen3.5-9B \citep{qwen3.5} to generate $n=5$ rephrased translations preserving comparable accuracy and fluency, using temperature $t = 1$ and top-$p = 0.95$.

To isolate length as the sole variable, we apply a two-stage quality
control procedure. First, we compute the reference-based MetricX score
between each generated candidate and the original reference. We use reference-based metrics as initial quality gates as they are substantially less susceptible to length bias than their QE counterparts (Appendix~\ref{appendix:reference_based_metrics_length_bias}), making them suitable quality gates for this purpose. Then, we select the two candidates with the best scores and compute the score difference. Candidates whose score differs by less than $\epsilon = 0.1$ are treated as near-duplicates offering insufficient variation, while those differing by more than $\delta = 1.0$ are treated as potential quality degradations; both cases are discarded.
The $\delta = 1.0$ threshold reflects the inherent subjectivity of segment-level evaluation: WMT metrics shared tasks have repeatedly shown that professional annotators frequently disagree on minor errors at the 1-point level \citep{deutsch-2025-wmt24, kocmi-2024-wmt-sharedtask}, so score differences within this range are as likely to reflect subjective variance as genuine quality differences. This automatic stage retained 1,851 pairs. Second, two professional bilingual annotators per language pair independently assessed both variants under MQM without being told which was designated shorter. We retained a pair only when both raters judged its variants equal in quality. This rejected 92 pairs and left 1,759; Appendix~\ref{appendix:human_rating} gives the instructions and per-language counts.

Ultimately, each source chunk is paired with two translations---one shorter and one longer. AutoMQM pairs for which either output cannot be parsed are excluded. For every valid pair, a shorter win, tie, and longer win receive preference credit of 1, 0.5, and 0, respectively. The table reports the macro-average across language pairs.

\begin{table*}[t!]
\centering
\small
\setlength{\tabcolsep}{3pt}
\begin{tabular}{l cccccccccc c}
\toprule
    \textbf{Metric} & \textbf{de} & \textbf{zh} & \textbf{es} &
    \textbf{fr} & \textbf{ja} & \textbf{ko} & \textbf{hi} &
    \textbf{ar} & \textbf{ru} & \textbf{pt} & \textbf{Avg} \\
\midrule
    \multicolumn{12}{l}{\textit{en $\rightarrow$ xx}} \\
    \cmidrule(lr){1-12}
    MetricX-24 QE
        & 47.3 & 51.9 & \good{60.6} & \good{60.9} & \good{56.8}
        & \good{61.5} & \good{55.8} & \good{63.7} & 47.5
        & \good{57.6} & \good{56.4} \\
    CometKiwi
        & \bad{41.8} & 45.5 & \good{56.4} & \good{57.5} & 51.5
        & \good{55.1} & \bad{44.2} & 52.7 & 53.8
        & 45.9 & 50.4 \\
    \addlinespace[2pt]
    xCOMET-XL
        & 48.4 & 53.2 & 52.1 & \good{58.6} & \bad{42.4}
        & 48.7 & \good{55.8} & \good{59.3} & \good{57.5}
        & 48.2 & 52.4 \\
    Few-shot AutoMQM (Llama-3.1-8B)
        & 49.0 & 52.6 & \bad{42.9} & \good{61.5} & \bad{44.4}
        & 50.0 & 45.2 & \good{56.9} & \bad{37.5}
        & \good{65.3} & 50.5 \\
    Few-shot AutoMQM (Qwen3-8B)
        & 54.9 & 50.0 & 50.0 & 52.0 & \bad{42.4}
        & 48.9 & 51.4 & 48.0 & 47.7
        & 53.6 & 49.9 \\
    Few-shot AutoMQM (Gemini-2.5-Flash)
        & \bad{36.8} & \bad{43.5} & \bad{37.2} & \bad{43.7}
        & \bad{43.9} & \bad{35.3} & \bad{43.7} & \bad{40.7}
        & \bad{39.4} & \bad{42.4} & \bad{40.6} \\
\midrule
    \multicolumn{12}{l}{\textit{xx $\rightarrow$ en}} \\
    \cmidrule(lr){1-12}
    MetricX-24 QE
        & \good{63.2} & \bad{44.0} & 53.0 & 53.7 & 47.1
        & \bad{44.9} & 49.1 & \good{58.5} & 51.7
        & 53.7 & 51.9 \\
    CometKiwi
        & \good{60.4} & \good{65.3} & \good{56.4} & \good{55.3}
        & \good{75.7} & \good{67.3} & \good{56.5} & 50.0
        & \good{60.7} & \good{55.8} & \good{60.4} \\
    \addlinespace[2pt]
    xCOMET-XL
        & \good{68.1} & 48.0 & 52.5 & \good{60.6} & 48.6
        & 50.0 & 46.3 & 47.9 & \good{56.2}
        & \good{56.8} & 53.5 \\
    Few-shot AutoMQM (Llama-3.1-8B)
        & 46.9 & 48.1 & \bad{43.5} & \good{56.7} & \good{60.9}
        & 48.9 & \good{55.6} & 48.3 & \good{56.0}
        & 48.8 & 51.4 \\
    Few-shot AutoMQM (Qwen3-8B)
        & \bad{42.5} & \good{56.0} & 47.7 & 48.2 & \bad{44.1}
        & 54.2 & 46.9 & 50.0 & 52.3
        & 46.8 & 48.9 \\
    Few-shot AutoMQM (Gemini-2.5-Flash)
        & 50.0 & \bad{39.3} & \bad{44.6} & \bad{42.0}
        & \bad{37.1} & 49.0 & \bad{40.7} & \bad{34.6}
        & \bad{39.9} & \bad{44.7} & \bad{42.2} \\
\bottomrule
\end{tabular}
\caption{
Preference for shorter translations (\%) by metric and language pair.
An unbiased metric should show 50\%.
\good{Green} cells indicate bias toward shorter translations, while
\bad{red} cells indicate bias toward longer translations. Ties receive half credit; Avg is the macro-average over the ten language pairs.
}
\label{tab:trans_length_pref}
\end{table*}

\paragraph{Results.} Table~\ref{tab:trans_length_pref} reveals metric-class-dependent behavior rather than a universal shorter preference. The learned QE metrics generally favor shorter candidates: the en$\rightarrow$xx/xx$\rightarrow$en macro-averages (\%) are 56.4/51.9 for \metricx, 50.4/60.4 for \cometkiwi, and 52.4/53.5 for xCOMET-XL. In contrast, the open-weight AutoMQM judges are close to the 50–50 baseline (50.5/51.4 for Llama-3.1-8B and 49.9/48.9 for Qwen3-8B), while Gemini-2.5-Flash consistently favors the longer candidate (40.6/42.2 shorter preference). Thus, the cumulative-passage experiment in Section~\ref{main-experiment_data-filtering} and this matched-paraphrase experiment probe distinct axes: increasing content length elicits penalty inflation across metrics, whereas changing verbosity for fixed content produces model-dependent preferences.

\paragraph{Tie and parse robustness.} Drop-ties and half-credit estimates differ by at most 0.1 percentage points for Llama and Qwen, indicating that their near-neutral results are not artifacts of tie handling. Parsing retains 731 and 940 pairs for Llama and Qwen, respectively, and failure rates increase with absolute candidate length. However, among pairs for which exactly one candidate fails to parse, we find no evidence of directional asymmetry (Llama: 297 shorter-side vs. 289 longer-side failures, two-sided exact binomial \(p=0.77\); Qwen: 259 vs. 275, \(p=0.52\)). The preference estimates therefore apply to successfully parsed pairs, which are somewhat shorter on average than the excluded pairs; Appendix~\ref{appendix:tie_parse_robustness} provides the full audit.

\paragraph{Natural expansion.} Because translations naturally expand or contract at different rates across language pairs and directions, the observed shorter-candidate preferences could reflect language-specific expected length ratios rather than length alone. To test whether natural expansion plays such a role, we estimate each language pair's mean target/source token ratio from WMT24++ references and correlate it with shorter-candidate preference across the ten language pairs, separately for each metric and direction. The correlations vary in sign and magnitude. Only the Pearson correlation for CometKiwi xx$\rightarrow$en is nominally significant ($r=0.686$, $p=0.028$); the corresponding Spearman correlation is not significant, and the pattern does not hold for other metrics. Thus, we find no consistent evidence that language-pair-specific natural expansion explains the observed preferences; Appendix~\ref{appendix:natural_expansion} gives the full analysis.

\section{Analysis}

The experiments in Section~\ref{sec:experiments} show pervasive penalty inflation as correct content accumulates, while matched-paraphrase preferences vary by model class. We now investigate the cumulative-length penalty in MetricX-24 QE, presenting three complementary pieces of evidence: the training distribution is skewed against long error-free examples (\S\ref{analysis:distribution}); increasing model capacity does not reliably reduce the effect (\S\ref{analysis:model_size}); and switching from raw error scores to error density at training time substantially reduces it (\S\ref{analysis:normalization}). Together, these results identify skewed supervision, rather than underfitting, as the primary source of MetricX's bias.

\subsection{Distribution of MQM Scores in Training Data}
\label{analysis:distribution}

\begin{figure*}[t!]
\centering
\begin{tabular}{c}
\includegraphics[width=\textwidth]{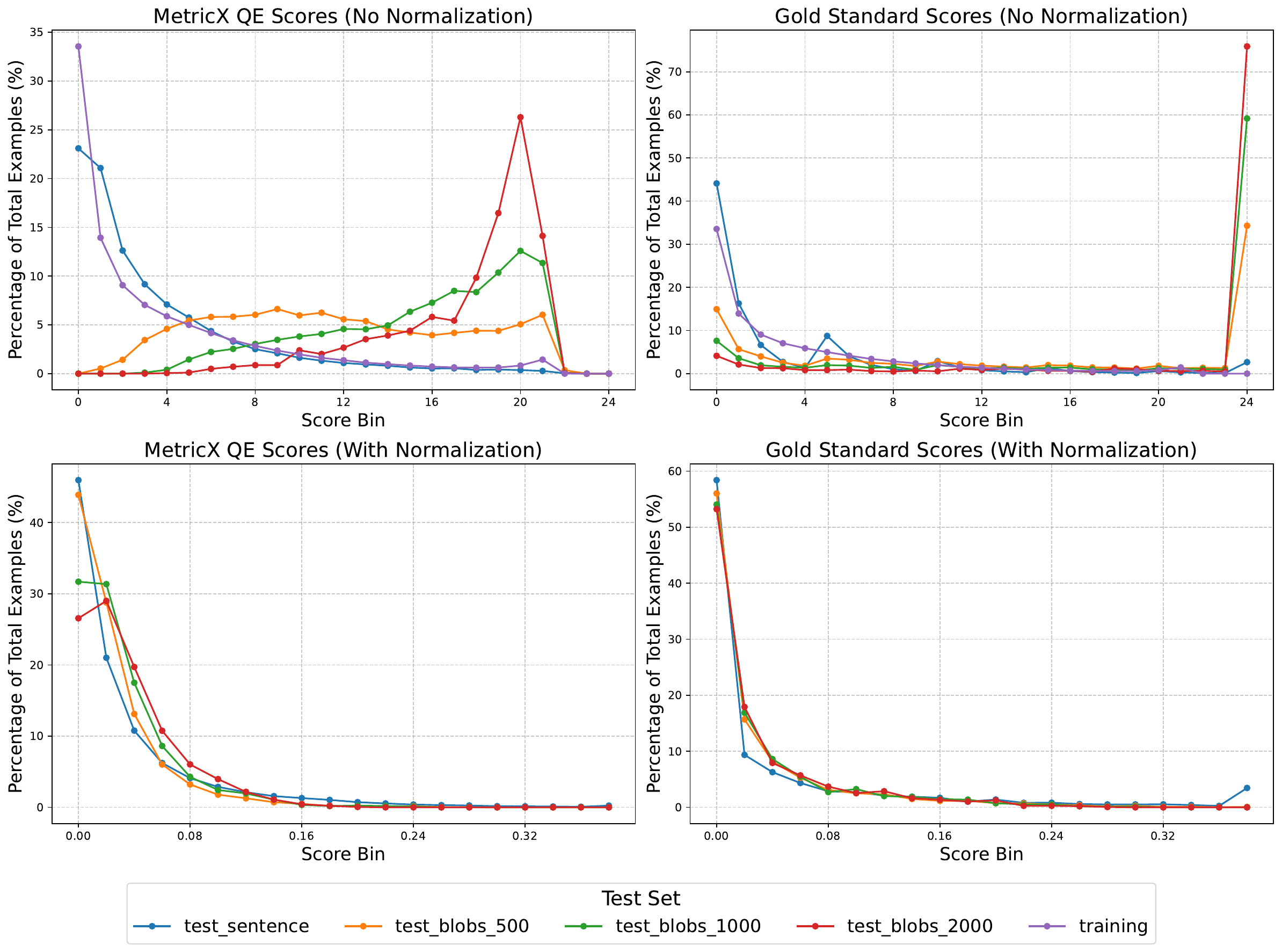}
\end{tabular}
\caption{Distribution of MetricX QE outputs and gold-standard scores. Top: without normalization. Bottom: with length normalization at training time. Normalization restores alignment with the ground-truth distribution across translations of varying length.}
\label{fig:distribution_of_mqm_scores}
\end{figure*}

We examine the distribution of gold-standard MQM scores in the training data used by learned QE models. We construct blob-level MQM test sets from WMT data, where a blob is a contiguous block of source--translation segments concatenated into a single evaluation instance, with a test-set-specific token limit of 500, 1k, or 2k. Figure~\ref{fig:distribution_of_mqm_scores} presents the distribution of gold-standard scores and \metricx's output scores on these test sets. The distribution of output scores is skewed relative to the gold standard, and as the number of input tokens increases, the proportion of perfect translations (gold score $= 0$) in the data diminishes. The output distribution mirrors this pattern, with the frequency of zero-score predictions dropping accordingly. The model has learned to associate sequence length with error probability, reflecting the distributional properties of its training data rather than a genuine relationship between length and quality.

\subsection{Effect of Model Capacity}
\label{analysis:model_size}

One hypothesis is that the bias stems from underfitting: smaller models may lack the capacity to disentangle length from quality in skewed training distributions. If so, scaling to larger models should progressively reduce the bias. We test this using the three publicly available MetricX-24 variants (Large, XL, and XXL) and repeat the translation length preference experiment from Section~\ref{translation_length_bias}.

\begin{table*}[t!]
\centering
\resizebox{\textwidth}{!}{
\begin{tabular}{ll cccccccccc c}
\toprule
    \textbf{Direction} & \textbf{Model} & \textbf{de} & \textbf{zh} & \textbf{es} & \textbf{fr} & \textbf{ja} & \textbf{ko} & \textbf{hi} & \textbf{ar} & \textbf{ru} & \textbf{pt} & \textbf{Avg} \\
\midrule
    \multirow{3}{*}{en $\rightarrow$ xx}
    & Large & \good{58.2} & \good{64.9} & 48.9 & \good{56.3} & \bad{39.4} & \bad{43.6} & \good{56.8} & \good{58.2} & \bad{40.0} & 49.4 & 51.6 \\
    & XL & \bad{44.0} & 51.9 & \good{57.4} & \good{58.6} & 51.5 & \good{59.0} & 52.6 & \good{60.4} & 48.8 & 54.1 & 53.8 \\
    & XXL & 49.5 & \good{57.1} & \good{55.3} & 50.6 & 45.5 & \good{61.5} & 54.7 & \good{63.7} & 51.2 & \good{57.6} & 54.7 \\
\midrule
    \multirow{3}{*}{xx $\rightarrow$ en}
    & Large & 49.5 & \bad{44.0} & 49.5 & \bad{43.6} & 45.7 & 54.1 & 49.1 & 53.2 & 52.8 & 54.7 & 49.6 \\
    & XL & \good{63.7} & \bad{42.7} & 50.5 & 50.0 & 48.6 & \bad{42.9} & 48.1 & \good{58.5} & 48.3 & 51.6 & 50.5 \\
    & XXL & 52.7 & 54.7 & 49.5 & 48.9 & 45.7 & \good{56.1} & \good{56.5} & 48.9 & 53.9 & \bad{44.2} & 51.1 \\
\bottomrule
\end{tabular}
}
\caption{Effect of MetricX model size on preference for shorter translations (\%) by translation direction. An unbiased metric should show 50.0. \good{Green} cells indicate bias toward shorter translations, while \bad{red} cells indicate bias toward longer translations. Colored cells deviate from 50.0 by more than 5 percentage points.}
\label{tab:ablation_size}
\end{table*}

Table~\ref{tab:ablation_size} shows that increasing model capacity does not reliably reduce length bias. The average short-translation preference (\%) does not decrease with model size (en$\rightarrow$xx: Large 51.6, XL 53.8, XXL 54.7), and all three variants exhibit substantial per-language biases in both directions. Larger models shift \emph{which} languages are affected and \emph{which direction} the bias takes, but do not systematically reduce it. Since all three models are trained on the same data with the same distributional skew, the persistence of bias across scales suggests that the root cause lies in the training distribution itself, rather than in the model's capacity.

\subsection{Diagnostic Intervention: Length Normalization}
\label{analysis:normalization}

\begin{figure*}[t!]
\centering
\begin{tabular}{c}
\includegraphics[width=\textwidth]{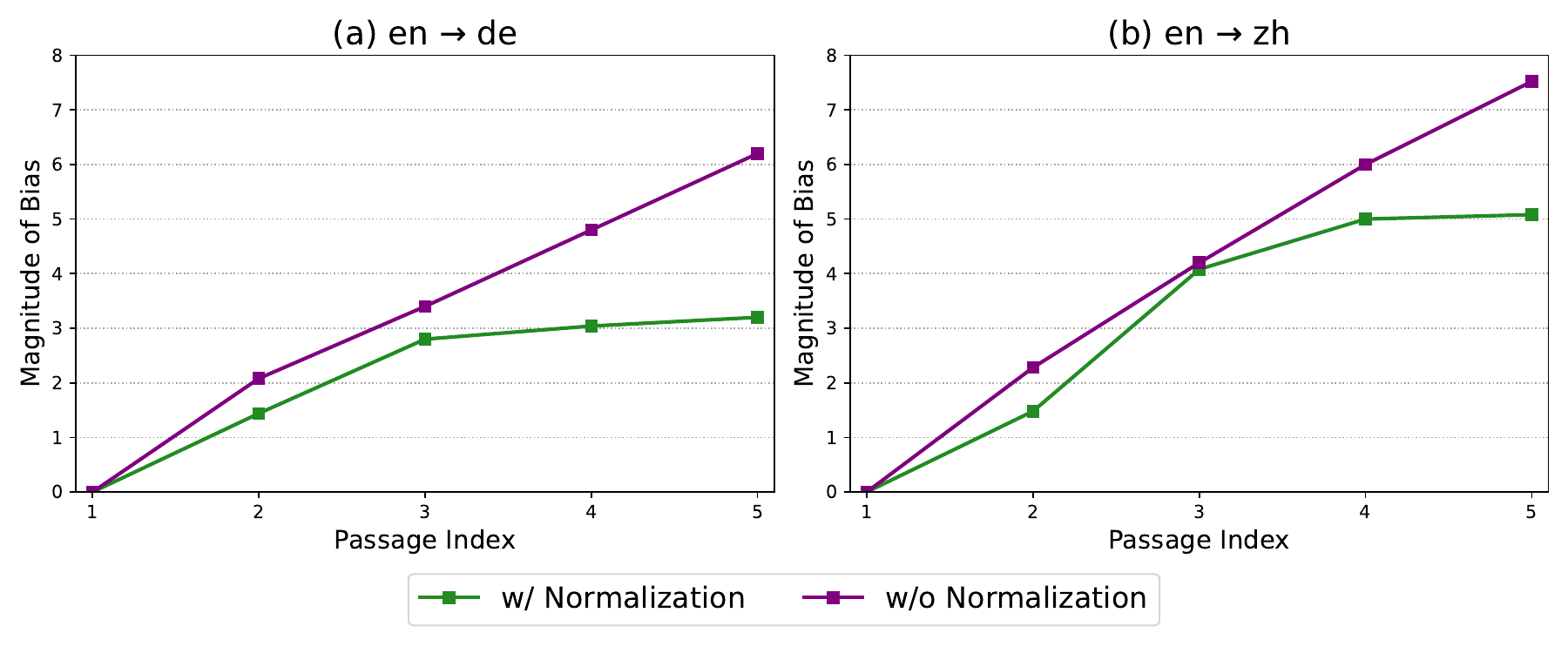}
\end{tabular}
\caption{Performance of \metricx with and without length normalization. We report the slope of score changes to show that length normalization mitigates length bias.}
\label{fig:normalization_at_training}
\end{figure*}
\vspace{-6pt}

Section~\ref{analysis:model_size} shows that scaling model capacity does not
resolve the bias, pointing to the training distribution as the root cause.
We now test this directly by applying length normalization at training time
as a diagnostic intervention. If correcting for the distributional skew
substantially reduces the bias, the hypothesis would be supported.

With normalization, the model is trained to predict the \textbf{error
density} of a hypothesis translation, defined as $\textrm{Density}(\hat{\theta}, l) = \hat{\theta}(x, y)/l$,
where $\hat{\theta}(x, y)$ is the error rating of hypothesis $y$ for source $x$ and
$l$ denotes its segment length. Because error density is largely invariant
to segment length in the gold-standard annotations, this reformulation
removes the skew that the unnormalized objective inherits. At inference time, predictions are rescaled back to error ratings via
$\hat{\theta}(x, y) = \textrm{Density}(\hat{\theta}, l) \cdot l$ for comparison with the unnormalized
baseline. Error density is therefore only a training-time diagnostic rather than a new definition of translation quality; human MQM ratings remain the quality target. On standard WMT meta-evaluation, the density-trained model remains within 1.5 points of the unnormalized baseline on SPA and Acc-23 across sentence and blob granularities (Appendix~\ref{appendix:error_density_meta_eval}).

We train a MetricX QE model with this normalization.
Figure~\ref{fig:distribution_of_mqm_scores} (bottom-left) shows the
resulting output distribution: the curves for different lengths collapse onto
a single, invariant density function, mirroring the ground-truth human
distribution (Figure~\ref{fig:distribution_of_mqm_scores}, bottom-right).
The unnormalized baseline fails to capture this property.

We further replicate the concatenation experiment from Section~\ref{main-experiment_data-filtering} and report results in Figure~\ref{fig:normalization_at_training}. Normalization reduces the magnitude of length bias beyond three segments, consistent with the distributional analysis above.
\vspace{-6pt}

\section{Conclusion}
In this work, we show that reference-free QE models are not length-invariant when assessing long-form translations: they inflate predicted error as translations grow longer, while matched-paraphrase comparisons reveal model-dependent behavior---learned QE metrics tend to favor shorter candidates, whereas LLM-as-a-Judge metrics range from being approximately length-neutral to favoring longer candidates. We trace MetricX-24 QE's cumulative-length penalty to a training pipeline artifact: the under-representation of long, error-free examples in supervision data. Scaling model capacity does not reliably reduce the bias, ruling out underfitting as the explanation. Instead, switching from raw error scores to error density at training time, which corrects for the distributional skew, substantially reduces the bias and restores closer alignment with human ground-truth distributions, supporting the conclusion that the bias is distributional in origin. While aggregate benchmarks often miss these tail-end failures, our findings highlight the necessity of auditing length invariance to ensure reliable evaluation in long-context applications.

\bibliography{main}
\bibliographystyle{colm2026_conference}

\appendix
\appendix
\section{Appendix}

\subsection{Length Bias in Reference-Based Metrics}
\label{appendix:reference_based_metrics_length_bias}
\begin{figure}[ht!]
\centering
\begin{tabular}{c}
\includegraphics[width=0.96\columnwidth]{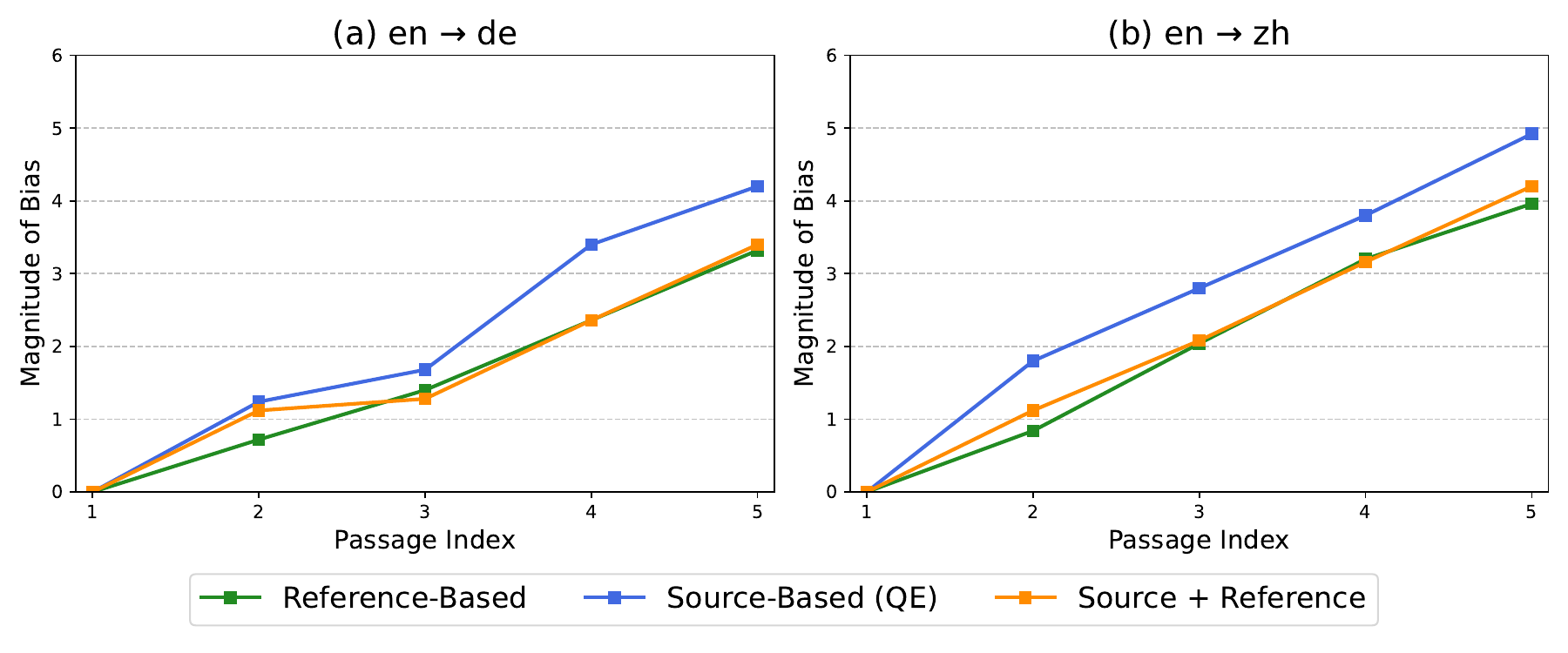}
\end{tabular}
\caption{Performance of three classes of MetricX-24: reference-based, source-based (QE), and hybrid. We report the slope of score changes to illustrate that metrics with access to references exhibit reduced length bias, though the effect nonetheless persists.}
\label{fig:reference-based_metrics}
\end{figure}

We compare reference-based, source-based (QE), and hybrid versions of MetricX-24 under the experimental setup described in Section~\ref{main-experiment_data-filtering}. Figure~\ref{fig:reference-based_metrics} shows that metrics with access to references exhibit attenuated length bias, plausibly because the reference translation provides an implicit prior on the expected length of a well-formed hypothesis. Nonetheless, residual length bias persists, indicating that reference access alone does not fully eliminate the effect.

\subsection{Human Rating and Quality Control}
\label{appendix:human_rating}

Two professional bilingual annotators were assigned to each language pair through a translation vendor and were paid \$55 USD/hour. For every pair, each annotator saw the source and both candidate translations, without shorter/longer labels, and assessed their accuracy and fluency under MQM. Annotators were instructed to decide whether the variants had equal overall quality; a pair was retained only when both independently judged them equal. Of 1,851 pairs passing automatic quality control, 92 were rejected at this stage, leaving the 1,759 pairs summarized in Table~\ref{tab:human_rating_counts}.

\begin{table}[ht!]
\centering
\small
\begin{tabular}{lrrr}
\toprule
\textbf{Language} & \textbf{en$\to$xx} & \textbf{xx$\to$en} & \textbf{Total} \\
\midrule
German (de\_DE)    & 91 & 91  & 182 \\
Chinese (zh\_CN)   & 77 & 75  & 152 \\
Spanish (es\_MX)   & 94 & 101 & 195 \\
French (fr\_FR)    & 87 & 94  & 181 \\
Japanese (ja\_JP)  & 66 & 70  & 136 \\
Korean (ko\_KR)    & 78 & 98  & 176 \\
Hindi (hi\_IN)     & 95 & 108 & 203 \\
Arabic (ar\_EG)    & 91 & 94  & 185 \\
Russian (ru\_RU)   & 80 & 89  & 169 \\
Portuguese (pt\_BR)& 85 & 95  & 180 \\
\midrule
Total      & 844 & 915 & 1,759 \\
\bottomrule
\end{tabular}
\caption{Human-validated translation pairs retained for the translation-length experiment.}
\label{tab:human_rating_counts}
\end{table}

\subsection{Full Results for Translation Length Preference Experiments}
\label{appendix:length_bias_at_trans_full_tables}

Tables~\ref{tab:appendix_metricx-xl_combined} and \ref{tab:appendix_cometkiwi_combined} present the full per-language breakdown of shorter-translation preference rates across all token count difference bins for MetricX-24 QE and CometKiwi, respectively, complementing the aggregated results of the translation length preference experiments in Section~\ref{translation_length_bias}. Ties receive half credit in all cells.

\begin{table*}[t!]
\centering
\resizebox{\textwidth}{!}{%
\begin{tabular}{l@{\hskip 4pt} cccc cccc}
\toprule
    & \multicolumn{4}{c}{\textit{en $\rightarrow$ xx}} & \multicolumn{4}{c}{\textit{xx $\rightarrow$ en}} \\
    \cmidrule(lr){2-5} \cmidrule(lr){6-9}
    \textbf{Lang} & \textbf{0--20} & \textbf{20--40} & \textbf{40--60} & \textbf{60+} & \textbf{0--20} & \textbf{20--40} & \textbf{40--60} & \textbf{60+} \\
\midrule
    de & 38.2 (55) & 60.9 (23) & 80.0 (5) & 50.0 (8) & 61.5 (61) & 75.0 (16) & 50.0 (8) & 66.7 (6) \\
    zh & 52.4 (42) & 59.1 (22) & 45.5 (11) & -- & 44.4 (36) & 35.0 (20) & 61.5 (13) & 33.3 (6) \\
    es & 63.2 (57) & 56.5 (23) & 50.0 (10) & -- & 51.4 (72) & 65.0 (20) & 8.3 (6) & -- \\
    fr & 58.1 (43) & 65.2 (23) & 58.3 (12) & 66.7 (9) & 57.7 (65) & 57.9 (19) & 25.0 (8) & -- \\
    ja & 47.9 (47) & 60.0 (10) & 100.0 (9) & -- & 50.0 (36) & 35.7 (14) & 40.0 (10) & 60.0 (10) \\
    ko & 61.2 (49) & 52.4 (21) & 83.3 (6) & -- & 38.5 (65) & 64.0 (25) & 37.5 (8) & -- \\
    hi & 47.2 (53) & 66.7 (18) & 61.5 (13) & 72.7 (11) & 45.5 (66) & 50.0 (32) & 83.3 (6) & -- \\
    ar & 69.0 (42) & 57.1 (21) & 53.3 (15) & 69.2 (13) & 53.6 (56) & 60.0 (20) & 58.3 (12) & 100.0 (6) \\
    ru & 43.6 (39) & 57.1 (14) & 57.1 (14) & 38.5 (13) & 49.0 (49) & 51.5 (33) & 66.7 (6) & -- \\
    pt & 56.4 (55) & 50.0 (16) & 63.6 (11) & -- & 60.3 (63) & 47.6 (21) & 30.0 (10) & -- \\
\bottomrule
\end{tabular}%
}
\caption{Preference for shorter translations (\%) by token count difference, MetricX-24 QE. Each column contains instances where the longer candidate exceeds the shorter by the stated token count range. Cells with fewer than 5 instances are shown as --. Count of instances shown in parentheses.}
\label{tab:appendix_metricx-xl_combined}
\end{table*}

\begin{table*}[t!]
\centering
\resizebox{\textwidth}{!}{%
\begin{tabular}{l@{\hskip 4pt} cccc cccc}
\toprule
    & \multicolumn{4}{c}{\textit{en $\rightarrow$ xx}} & \multicolumn{4}{c}{\textit{xx $\rightarrow$ en}} \\
    \cmidrule(lr){2-5} \cmidrule(lr){6-9}
    \textbf{Lang} & \textbf{0--20} & \textbf{20--40} & \textbf{40--60} & \textbf{60+} & \textbf{0--20} & \textbf{20--40} & \textbf{40--60} & \textbf{60+} \\
\midrule
    de & 43.6 (55) & 39.1 (23) & 80.0 (5) & 12.5 (8) & 57.4 (61) & 62.5 (16) & 75.0 (8) & 66.7 (6) \\
    zh & 38.1 (42) & 54.5 (22) & 54.5 (11) & -- & 58.3 (36) & 60.0 (20) & 76.9 (13) & 100.0 (6) \\
    es & 57.9 (57) & 56.5 (23) & 50.0 (10) & -- & 52.8 (72) & 60.0 (20) & 66.7 (6) & -- \\
    fr & 58.1 (43) & 52.2 (23) & 66.7 (12) & 55.6 (9) & 50.8 (65) & 63.2 (19) & 62.5 (8) & -- \\
    ja & 51.1 (47) & 50.0 (10) & 55.6 (9) & -- & 61.1 (36) & 85.7 (14) & 90.0 (10) & 100.0 (10) \\
    ko & 59.2 (49) & 42.9 (21) & 66.7 (6) & -- & 55.4 (65) & 92.0 (25) & 87.5 (8) & -- \\
    hi & 49.1 (53) & 16.7 (18) & 53.8 (13) & 54.5 (11) & 48.5 (66) & 75.0 (32) & 33.3 (6) & -- \\
    ar & 45.2 (42) & 52.4 (21) & 60.0 (15) & 69.2 (13) & 46.4 (56) & 50.0 (20) & 58.3 (12) & 66.7 (6) \\
    ru & 61.5 (39) & 64.3 (14) & 50.0 (14) & 23.1 (13) & 55.1 (49) & 63.6 (33) & 83.3 (6) & -- \\
    pt & 41.8 (55) & 43.8 (16) & 54.5 (11) & -- & 47.6 (63) & 71.4 (21) & 70.0 (10) & -- \\
\bottomrule
\end{tabular}%
}
\caption{Preference for shorter translations (\%) by token count difference, CometKiwi. Each column contains instances where the longer candidate exceeds the shorter by the stated token count range. Cells with fewer than 5 instances are shown as --. Count of instances shown in parentheses.}
\label{tab:appendix_cometkiwi_combined}
\end{table*}

\subsection{Tie Handling and Parse-Failure Robustness}
\label{appendix:tie_parse_robustness}

Table~\ref{tab:tie_parse_robustness} compares strict shorter wins, dropping ties, and assigning ties half credit after pooling all language pairs and directions. The strict rule treats every tie as a shorter loss and therefore understates shorter preference for discrete AutoMQM scores. Drop-ties and half-credit agree closely. For Llama and Qwen, parse failure is driven primarily by absolute candidate length but is direction-symmetric: among pairs for which exactly one side failed, the longer side did not fail more often ($p=0.77$ and $0.52$). The retained samples therefore show no directional attrition, although they underrepresent the longest inputs.

\begin{table*}[t!]
\centering
\small
\setlength{\tabcolsep}{4pt}
\begin{tabular}{lrrrrrr}
\toprule
\textbf{Metric} & \textbf{Valid (N)} & \textbf{Parse S/L (\%)} & \textbf{Ties (\%)} & \textbf{Strict (\%)} & \textbf{Drop ties (\%)} & \textbf{Half credit (\%)} \\
\midrule
MetricX-24 QE & 1,759 & -- & 0.5 & 54.0 & 54.2 & 54.2 \\
CometKiwi & 1,759 & -- & 0.0 & 55.3 & 55.3 & 55.3 \\
xCOMET-XL & 1,759 & -- & 0.0 & 53.2 & 53.2 & 53.2 \\
Llama-3.1-8B  & 731 & 58.0/58.4 & 14.5 & 43.5 & 50.9 & 50.8 \\
Qwen3-8B  & 940 & 69.1/68.2 & 18.2 & 40.4 & 49.4 & 49.5 \\
Gemini-2.5-Flash  & 1,759 & 100/100 & 11.7 & 35.7 & 40.4 & 41.5 \\
\bottomrule
\end{tabular}
\caption{Tie and parse-failure audit pooled across language pairs and translation directions. Parse S/L gives candidate-level parse success for the shorter/longer side; learned metrics do not require structured-output parsing.}
\label{tab:tie_parse_robustness}
\end{table*}

\subsection{Robustness to Natural Translation Expansion}
\label{appendix:natural_expansion}

Because translations naturally expand or contract at different rates across language pairs and directions, shorter-candidate preferences could reflect language-specific expected length ratios rather than length alone. For each language pair and direction, we compute the natural expansion ratio as the mean target/source token ratio over WMT24++ human gold references, using the MetricX tokenizer. We then compute each metric's Section~\ref{translation_length_bias} bias score as the mean shorter-candidate preference rate, assigning half credit to ties. Table~\ref{tab:natural_expansion} reports Spearman and Pearson correlations between expansion ratio and bias score across the ten language pairs, separately for each metric and direction. Correlations vary in sign and magnitude. Although the Pearson correlation for CometKiwi xx$\to$en is nominally significant, its Spearman correlation is not, and the pattern is not consistent across other metrics. The results therefore do not support natural translation expansion as the primary explanation for the observed shorter-candidate preferences.

\begin{table*}[t!]
\centering
\small
\setlength{\tabcolsep}{5pt}
\begin{tabular}{@{}llrrc@{}}
\toprule
\textbf{Metric} & \textbf{Direction} & \textbf{Spearman $\rho$ ($p$)} & \textbf{Pearson $r$ ($p$)} & \textbf{$N$} \\
\midrule
MetricX-24 QE & en$\to$xx & $+0.491$ ($0.150$) & $+0.212$ ($0.556$) & 10 \\
              & xx$\to$en & $-0.127$ ($0.726$) & $-0.276$ ($0.440$) & 10 \\
CometKiwi     & en$\to$xx & $+0.273$ ($0.446$) & $-0.066$ ($0.856$) & 10 \\
              & xx$\to$en & $+0.491$ ($0.150$) & $+0.686$ ($0.028$) & 10 \\
xCOMET-XL     & en$\to$xx & $+0.406$ ($0.244$) & $+0.469$ ($0.172$) & 10 \\
              & xx$\to$en & $+0.018$ ($0.960$) & $-0.128$ ($0.726$) & 10 \\
Llama-3.1-8B AutoMQM & en$\to$xx & $+0.370$ ($0.293$) & $+0.067$ ($0.854$) & 10 \\
              & xx$\to$en & $-0.139$ ($0.701$) & $+0.113$ ($0.756$) & 10 \\
Qwen3-8B AutoMQM & en$\to$xx & $+0.365$ ($0.300$) & $+0.463$ ($0.178$) & 10 \\
              & xx$\to$en & $-0.079$ ($0.829$) & $+0.079$ ($0.829$) & 10 \\
Gemini-2.5-Flash AutoMQM & en$\to$xx & $+0.006$ ($0.987$) & $+0.000$ ($1.000$) & 10 \\
              & xx$\to$en & $-0.370$ ($0.293$) & $-0.306$ ($0.390$) & 10 \\
\bottomrule
\end{tabular}
\caption{Correlation between language-pair-level natural expansion ratio and shorter-candidate preference. Expansion is the mean target/source token ratio over WMT24++ human gold references; preference assigns half credit to ties. Correlations are computed across ten language pairs for each metric and direction; $p$-values are in parentheses.}
\label{tab:natural_expansion}
\end{table*}

\subsection{Standard Meta-Evaluation after Error-Density Training}
\label{appendix:error_density_meta_eval}

In our analysis, density normalization is used only to test whether length bias arises from the supervision signal (Sections~\ref{analysis:distribution} and~\ref{analysis:normalization}); it removes the length-conditioned skew during training, and predictions are rescaled to raw error ratings at inference.

To verify that this intervention does not harm standard evaluation quality, we evaluate the density-trained model on standard WMT meta-evaluation at sentence and blob granularities. Table~\ref{tab:error_density_meta_eval} shows that it remains within 1.5 points of the unnormalized baseline on both Soft Pairwise Accuracy (SPA) and Acc-23 across all granularities and closely matches the baseline at the sentence level. Thus, the reduction in length bias does not come at the cost of agreement with human judgments.

\begin{table*}[ht!]
\centering
\small
\setlength{\tabcolsep}{7pt}
\begin{tabular}{@{}lrrrr@{}}
\toprule
& \multicolumn{2}{c}{\textbf{SPA}} & \multicolumn{2}{c}{\textbf{Acc-23}} \\
\cmidrule(lr){2-3}\cmidrule(lr){4-5}
\textbf{Test data} & \textbf{Baseline} & \textbf{Density-trained} & \textbf{Baseline} & \textbf{Density-trained} \\
\midrule
Sentence   & 86.89 & 87.76 & 58.81 & 58.87 \\
Blob (500) & 87.22 & 87.66 & 66.38 & 65.45 \\
Blob (1,000) & 86.62 & 86.57 & 71.55 & 70.67 \\
Blob (2,000) & 85.86 & 84.70 & 74.53 & 73.09 \\
\bottomrule
\end{tabular}
\caption{Standard WMT meta-evaluation for the baseline and error-density-trained models. Parentheses in the blob rows give the token limit. Higher values are better.}
\label{tab:error_density_meta_eval}
\end{table*}

\subsection{Proportion of Scores with Decreasing Trends in Data Filtering}
\label{appendix:proportion_of_decreasing_trends_data_filtering}

We analyze the proportion of documents exhibiting a decreasing trend in translation scores. For each language pair, we compare the score of passage 1 (containing segment 1) with passage 5 (containing segments 1--5), and report (i) the total number of documents and (ii) the proportion displaying a decreasing trend. Results for \metricx are summarized in Table~\ref{tab:proportion_of_scores_with_decreasing_trends}. Approximately 80\% of documents across 10 languages show a decreasing trend.

\begin{table}[ht!]
\centering
\begin{tabular}{|l|c|c|}
\toprule
\textbf{Language} & \textbf{\# of Documents} & \textbf{Proportion of scores with decreasing trends (\%)}\\
\midrule
German (de\_DE) & 48 & 85.4 \\
Chinese (zh\_CN) & 53 & 79.2 \\
Spanish (es\_MX) & 47 & 74.5 \\
French (fr\_FR) & 45 & 82.2 \\
Japanese (ja\_JP) & 54 & 90.7 \\
Korean (ko\_KR) & 46 & 82.6 \\
Hindi (hi\_IN) & 37 & 75.7 \\
Arabic (ar\_EG) & 48 & 68.8 \\
Russian (ru\_RU) & 48 & 79.2 \\
Portuguese (pt\_BR) & 46 & 80.4 \\
\midrule
Total & 472 & 80.1 \\
\bottomrule
\end{tabular}
\caption{Proportion of scores with decreasing trends with \metricx}
\label{tab:proportion_of_scores_with_decreasing_trends}
\end{table}


\subsection{Error Type / Severity Perturbation Prompt}
\label{appendix:error_type_and_severity}
\begin{figure}[ht!]
\center
\begin{tabular}{c}
  \includegraphics[width=0.9\columnwidth]{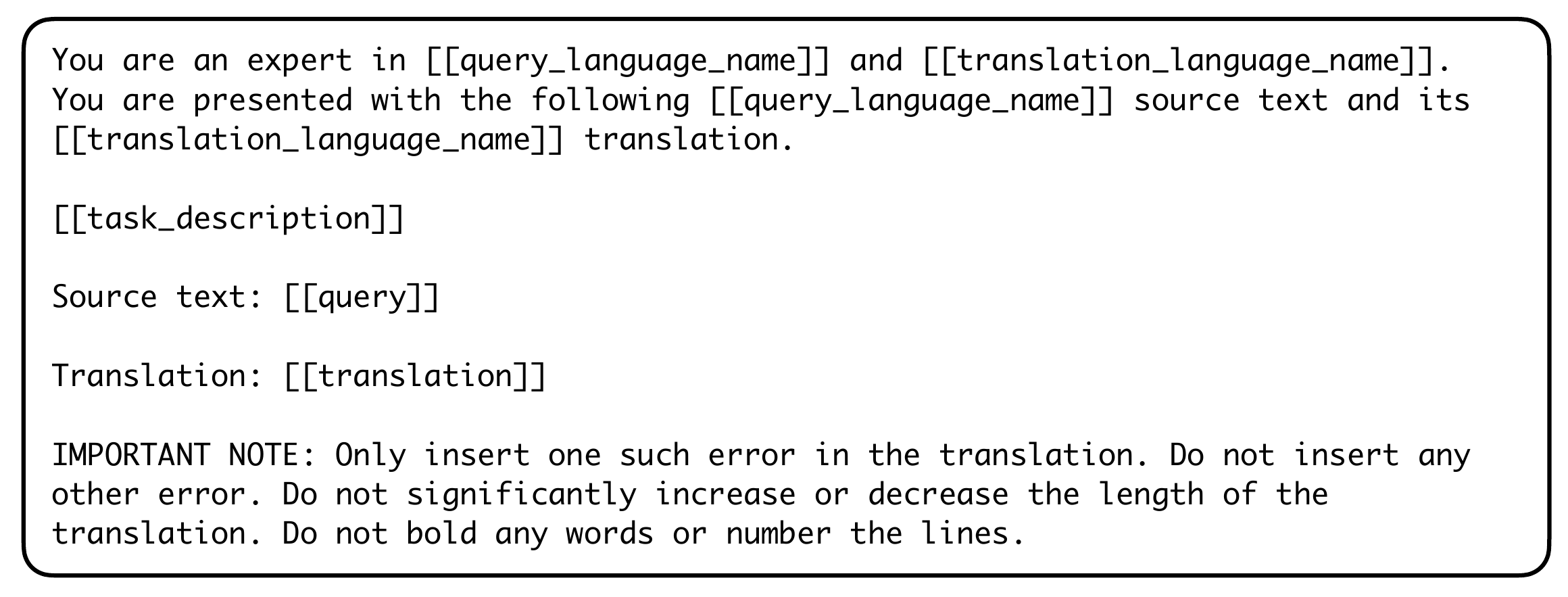} \\
\end{tabular}
\caption{Error perturbation prompt}
\label{fig:error_type_severity_prompt}
\end{figure}

\begin{figure}[ht!]
\center
\begin{tabular}{c}
  \includegraphics[width=0.9\columnwidth]{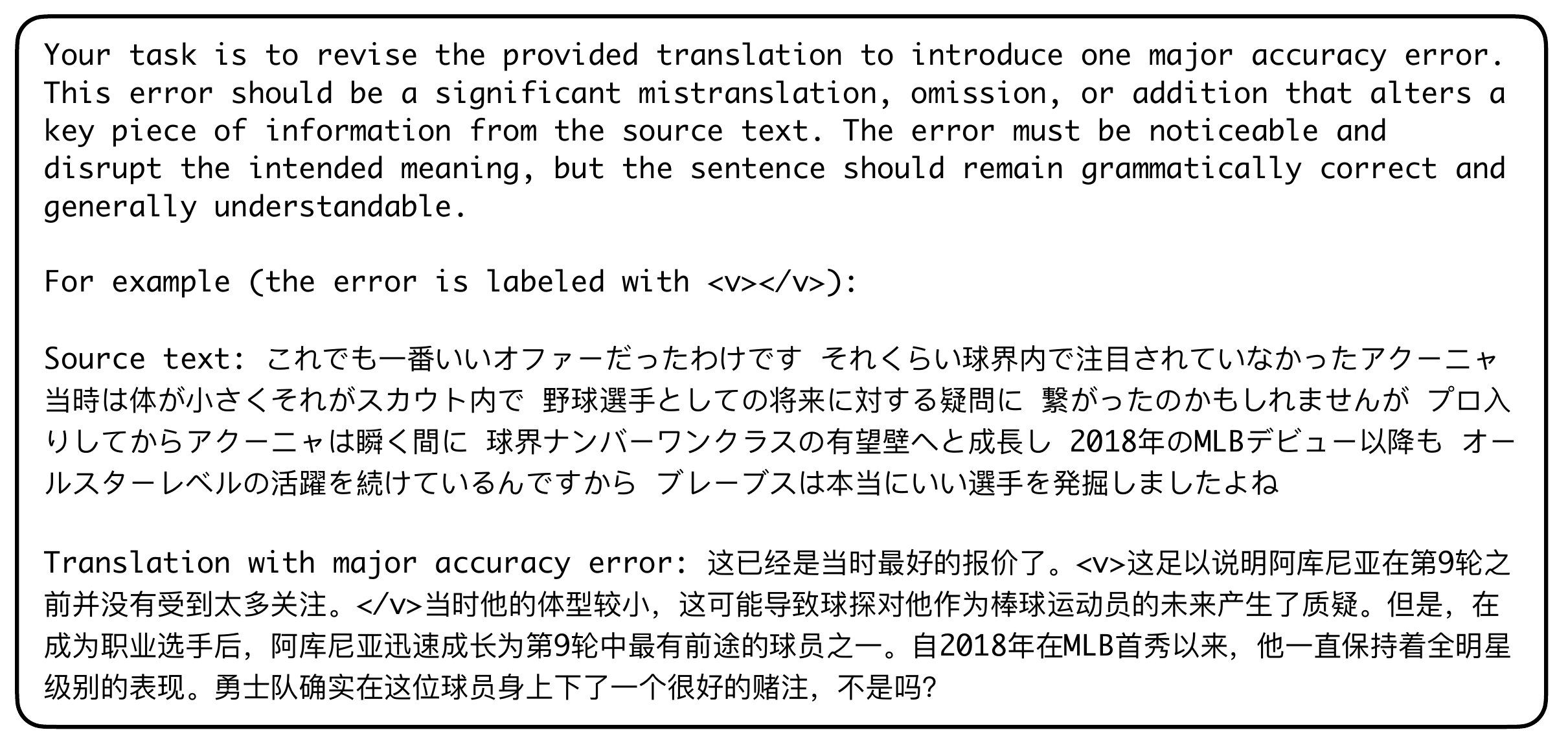} \\
\end{tabular}
\caption{Major accuracy prompt}
\label{fig:major_accuracy_prompt}
\end{figure}

\begin{figure}[ht!]
\center
\begin{tabular}{c}
  \includegraphics[width=0.9\columnwidth]{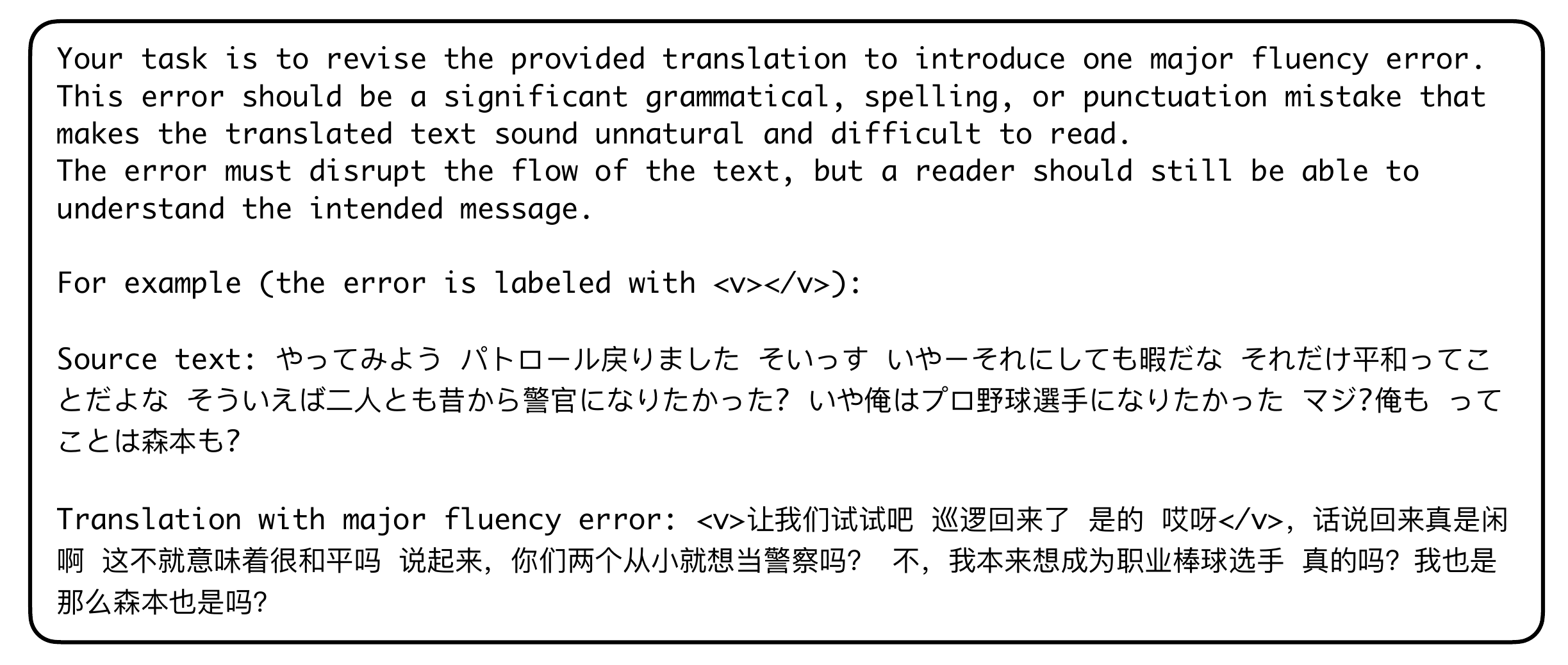} \\
\end{tabular}
\caption{Major fluency prompt}
\label{fig:major_fluency_prompt}
\end{figure}

\begin{figure}[ht!]
\center
\begin{tabular}{c}
  \includegraphics[width=0.9\columnwidth]{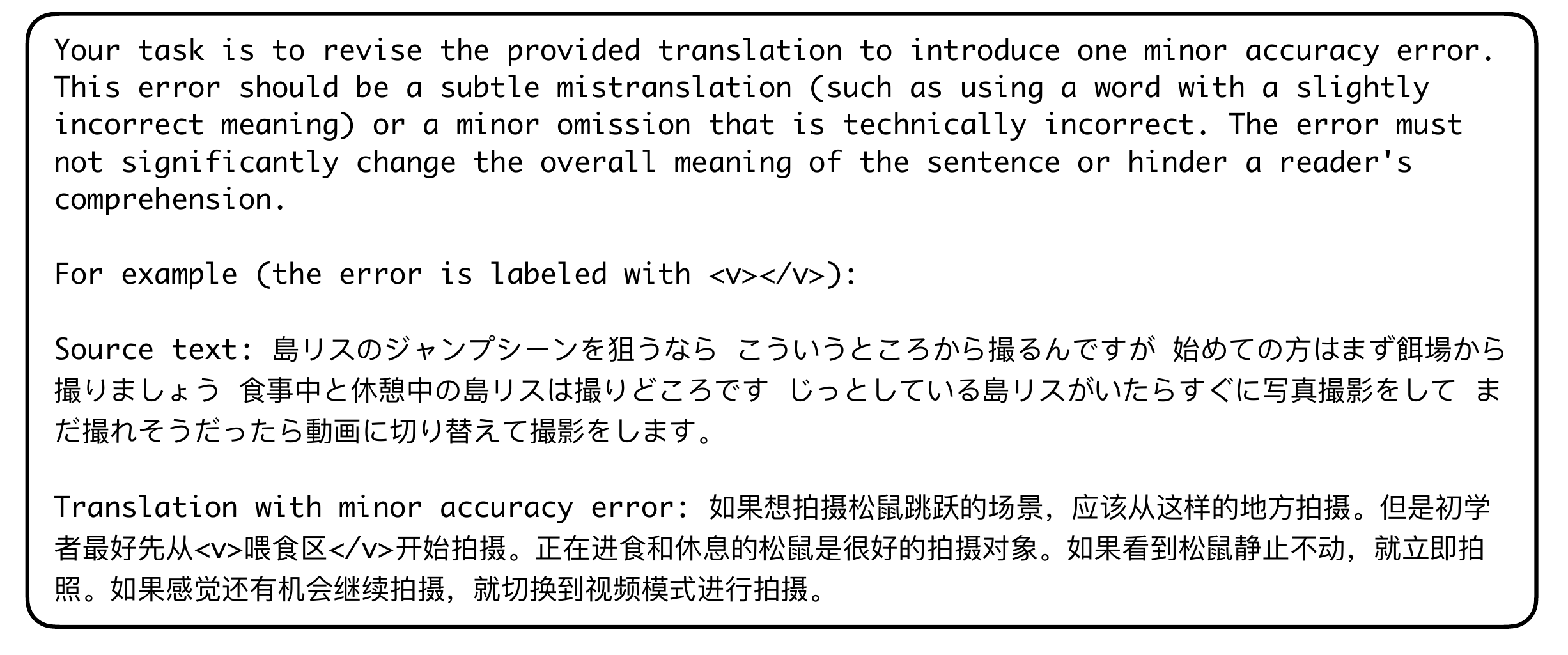} \\
\end{tabular}
\caption{Minor accuracy prompt}
\label{fig:minor_accuracy_prompt}
\end{figure}

\begin{figure}[ht!]
\center
\begin{tabular}{c}
  \includegraphics[width=0.9\columnwidth]{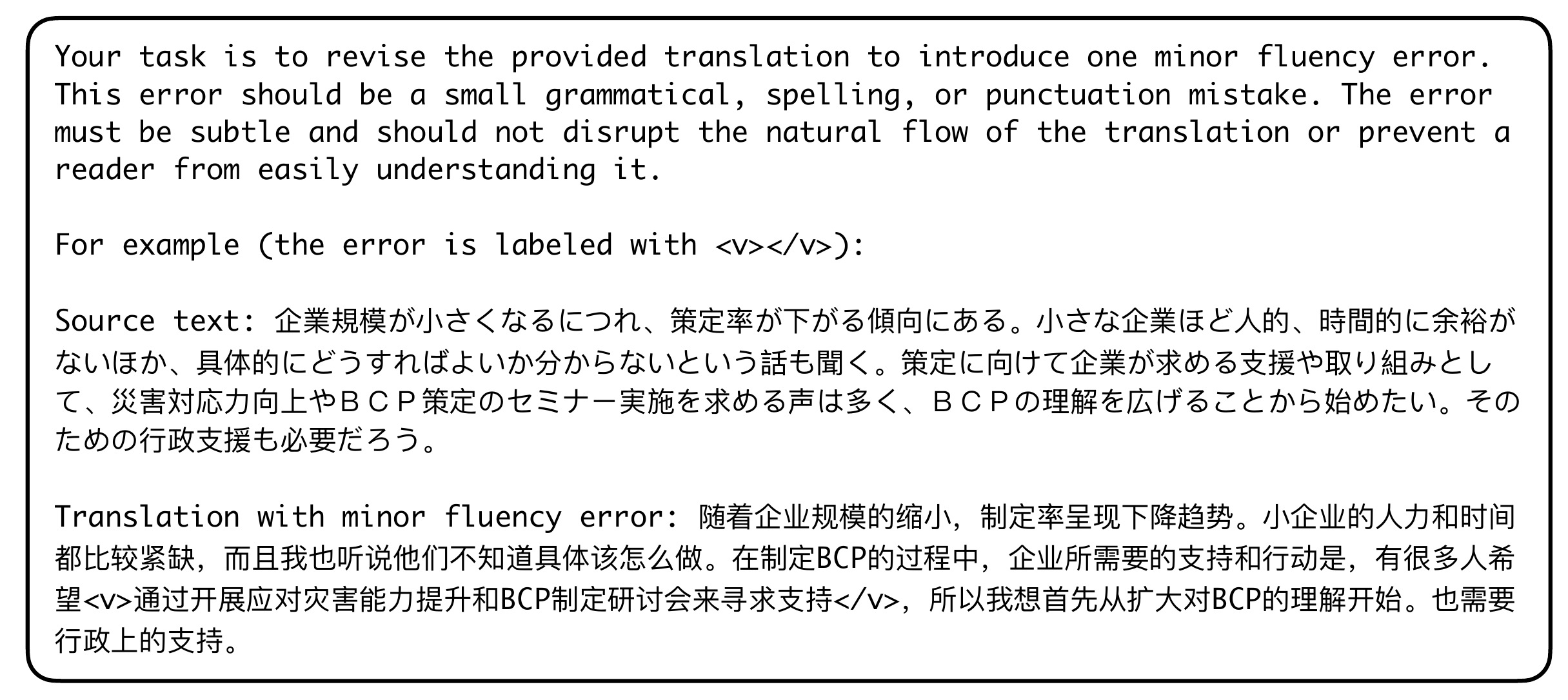} \\
\end{tabular}
\caption{Minor fluency prompt}
\label{fig:minor_fluency_prompt}
\end{figure}

Figures~\ref{fig:error_type_severity_prompt}--\ref{fig:minor_fluency_prompt} present the prompts used for error perturbation with different types and severity using Gemini-2.5-Pro.

\subsection{Few-shot AutoMQM Prompt}
\label{appendix:automqm_prompt}
\begin{figure}[ht!]
\center
\begin{tabular}{c}
  \includegraphics[width=0.8\columnwidth]{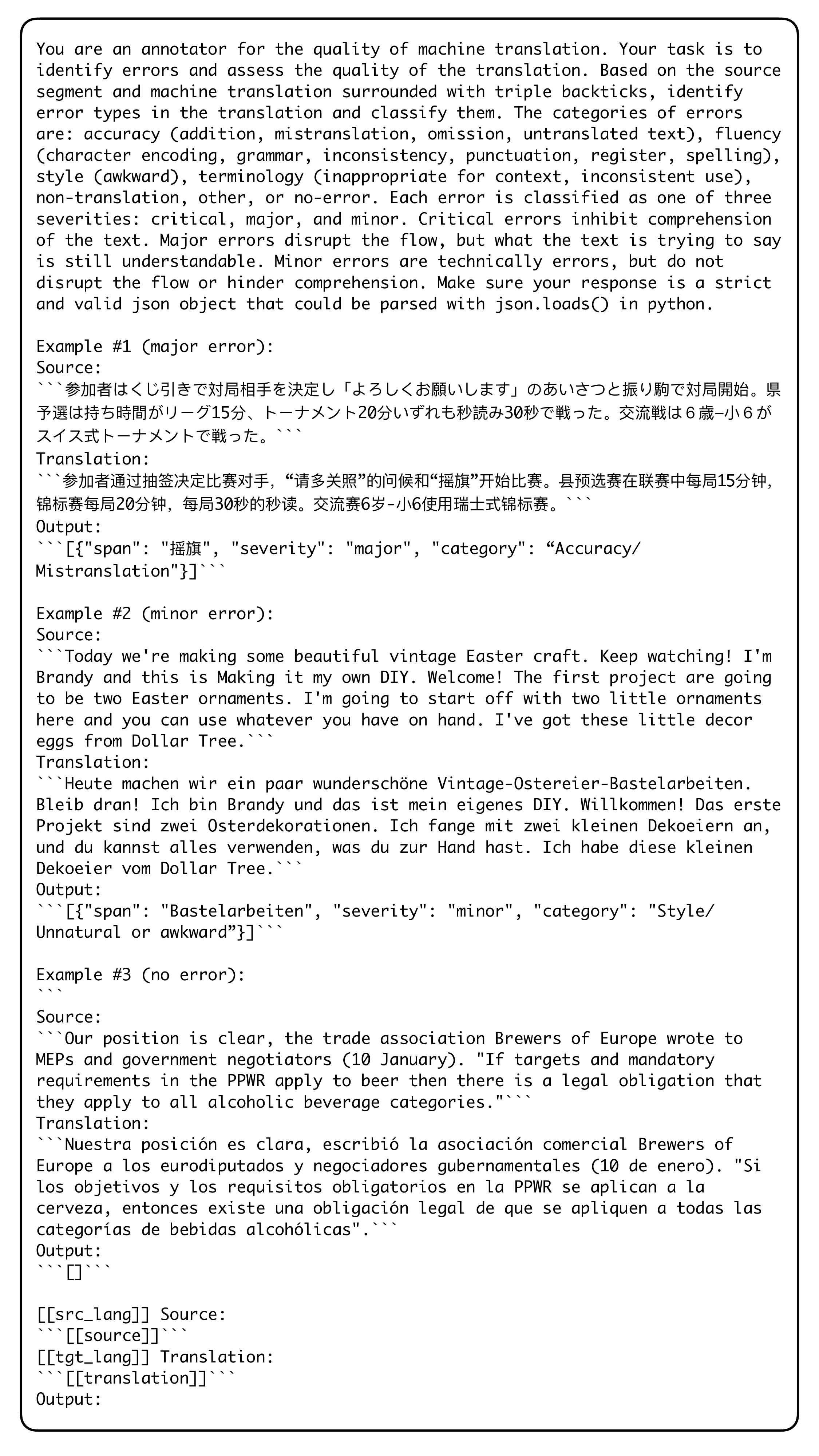} \\
\end{tabular}
\caption{Few-shot AutoMQM prompt}
\label{fig:automqm_prompt}
\end{figure}

Figure~\ref{fig:automqm_prompt} presents the shared prompt used by all few-shot AutoMQM evaluators: Llama-3.1-8B, Qwen3-8B, Gemini-2.5-Pro, and Gemini-2.5-Flash.

\end{document}